%% file: main.tex
\documentclass[10pt,twocolumn,letterpaper]{article}
\usepackage[pagenumbers]{cvpr}
\input{preamble}

\input{title}
\begin{document}
\maketitle
\input{sections/00_abstract}
\input{sections/01_introduction}
\input{sections/02_related_work}
\input{sections/03_study_design}
\input{sections/04_findings}
\input{sections/06_mechanism}
\input{sections/05_lite}
\input{sections/07_discussion}
\input{sections/08_conclusion}
\section*{Acknowledgments}
OpenAI Codex assisted with environment and training code, experiment preparation,
manuscript drafting and editing, and figure/table
generation code.
{\small
\bibliographystyle{ieeenat_fullname}
\bibliography{references}
}
\input{appendix}

\end{document}

%% file: preamble.tex
\usepackage{graphicx}
\usepackage{amsmath,amssymb}
\usepackage{booktabs,multirow}
\usepackage{xcolor,colortbl}
\usepackage{microtype}
\definecolor{cvprblue}{rgb}{0.21,0.49,0.74}
\usepackage[breaklinks,colorlinks,allcolors=cvprblue]{hyperref}
\usepackage[capitalize]{cleveref}
\newcommand{\gym}{MobileGym}
\newcommand{\lite}{MobileGym-Lite}
\newcommand{\fzero}{F0}
\newcommand{\fone}{F1}

\newcommand{\appref}[1]{Appendix~\ref{#1}}
\input{tables/numbers}

%% file: tables/numbers.tex
\newcommand{\FullSR}{51.61}
\newcommand{\FzeroSR}{56.13}
\newcommand{\LiteSR}{55.48}

%% file: title.tex
\newcommand{\papertitle}{How Complete Must an Environment Be to Train a Capable GUI Agent?}
\ifdefined\standalonesupplement
  \title{\papertitle\\[0.5em]Supplementary Material}
  \hypersetup{pdftitle={\papertitle\space (Supplementary Material)}}
\else
  \title{\papertitle}
  \hypersetup{pdftitle={\papertitle}}
\fi
\newcommand{\authoremail}[1]{\href{mailto:#1}{\nolinkurl{#1}}}
\author{%
  Jingxing Wang\textsuperscript{1}\quad Teng Wang\textsuperscript{2}\quad
  Xingyu Lou\textsuperscript{2}\quad Zihan Guo\textsuperscript{3}\\
  Jun Wang\textsuperscript{2,*}\quad Weiwen Liu\textsuperscript{1,*}\quad
  Weinan Zhang\textsuperscript{1,*}\quad Jianghao Lin\textsuperscript{1,*}\\[0.4em]
  {\normalsize \textsuperscript{1}Shanghai Jiao Tong University\quad
   \textsuperscript{2}OPPO Research Institute\quad
   \textsuperscript{3}Shanghai Innovation Institute}\\
  {\small \authoremail{01lightyear@sjtu.edu.cn}, \authoremail{wangjun7@oppo.com},
   \authoremail{wwliu@sjtu.edu.cn},}\\
  {\small \authoremail{wnzhang@sjtu.edu.cn}, \authoremail{linjianghao@sjtu.edu.cn}}\\
  {\normalsize Code: \url{https://github.com/Hydr0pon1c/Mobilegym-lite}}}
\hypersetup{pdfauthor={Jingxing Wang, Teng Wang, Xingyu Lou, Zihan Guo, Jun Wang, Weiwen Liu, Weinan Zhang, Jianghao Lin}}

%% file: sections/00_abstract.tex
\begin{abstract}
Graphical user interface (GUI) agents are widely trained in executable
GUI environments. To lower cost, build environments in batches, and
reduce manual effort, each work trades away part of the environment
through its construction choices.
We describe these choices along three dimensions: reward coverage, the
goal-reaching trajectories the verifier rewards; data coverage, the
application data the agent is exposed to beyond what the task reads or
writes; and feature coverage, the
application features beyond what the task uses. Existing works set each
dimension on engineering grounds, but they do not measure its effect on
training. How complete must an environment be to train a capable agent,
and along which dimensions?
In \gym{}, we adjust each dimension separately, train
Qwen3-VL-4B-Instruct, and evaluate every policy in the original full
environment.
A verifier that rewards only one reference path lowers success by 10.53
points, and removing unneeded records also lowers success. Pruning
unneeded features instead raises success from \FullSR\% to \FzeroSR\%,
entirely on tasks whose environments it left unchanged. The gain reverses
when each episode is weighted equally and vanishes when only episodes on the
pruned tasks are balanced. Without pruning, length weighting there
reinforces long detour successes that erode skills on other tasks.
Following these findings, we build \lite{}, which keeps reward and data
coverage complete and prunes unneeded features. It outperforms Full in
the full environment (\LiteSR\% versus \FullSR\% success) with lower
construction cost.
\end{abstract}

%% file: sections/01_introduction.tex
\section{Introduction}
\label{sec:intro}
Graphical user interface (GUI) agents are increasingly trained and
evaluated in executable GUI environments. Benchmarks such
as WebArena, VisualWebArena, AndroidWorld, and OSWorld check task
completion through
interaction~\cite{zhou2024webarena,koh2024visualwebarena,rawles2025androidworld,xie2024osworld},
and a growing line of work trains agents inside such environments with
reinforcement learning or synthesized
experience~\cite{bai2024digirl,qi2025webrl,wu2026mobilegym,cao2026guigenesis,sun2025osgenesis}.
Building these environments is costly, so each work makes its own
construction choices, trading part of the environment for lower cost,
batch automation, and less manual effort. Therefore, we ask: \emph{how complete must
a GUI environment be to train a capable GUI agent, and along which
dimensions?}

\begin{figure}[t]
  \centering
  \includegraphics[width=\columnwidth]{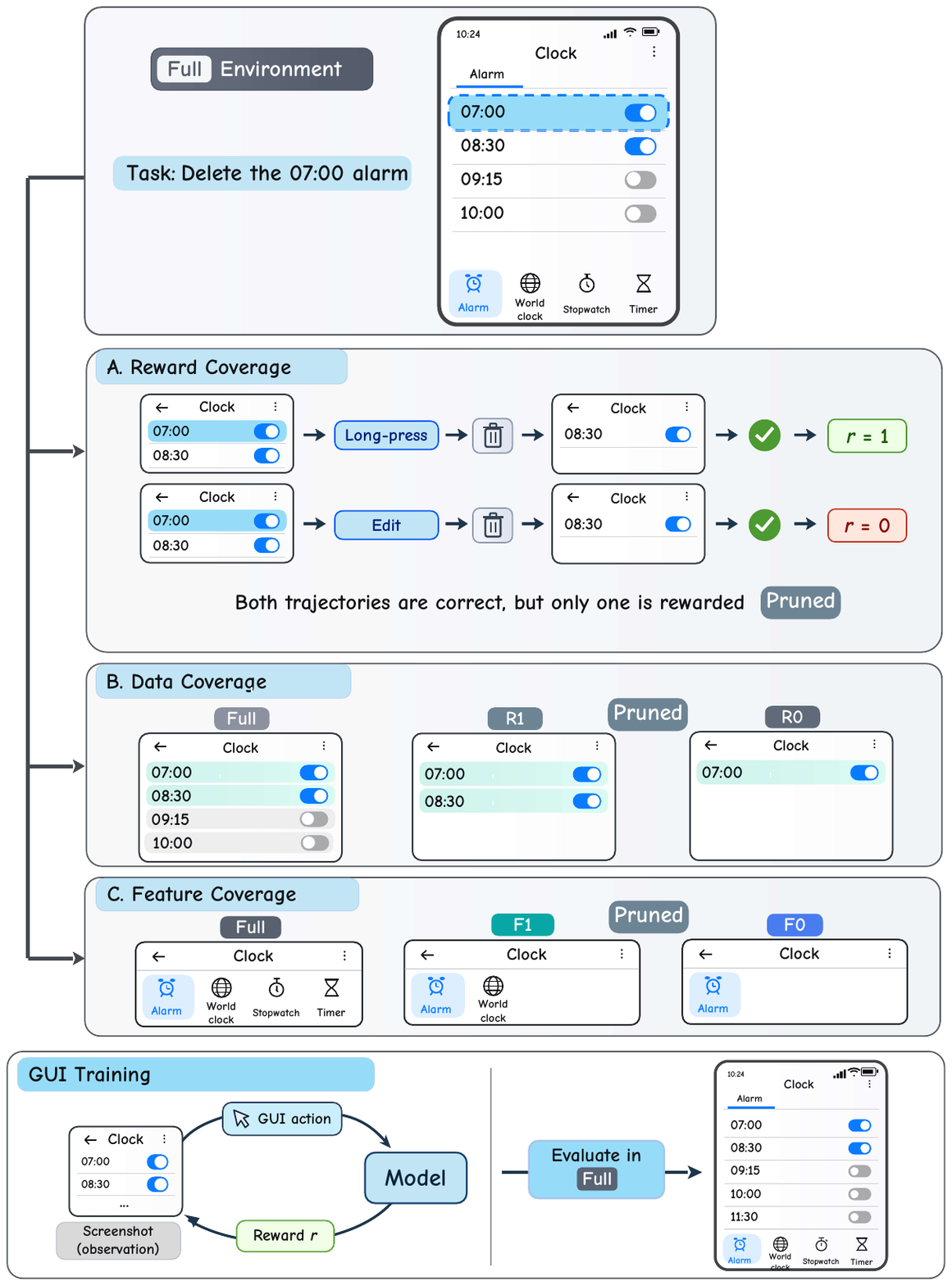}
  \caption{\textbf{Three dimensions of GUI environment construction.}
  Top: reward coverage. A verifier rewards long-press
  deletion but not an edit-sheet solution that reaches the same goal.
  Middle: data coverage. R0/R1/R2 retain the records the task needs
  and vary the unrelated ones. Bottom: feature coverage. Full/F0/F1
  expose different application functions. Policies train in the reduced
  environments and are evaluated in Full.}
  \label{fig:overview}
\end{figure}

We describe construction choices along three dimensions
(\cref{fig:overview}). Viewing the environment as a partially observable
Markov decision process, the builder writes a verifier that defines the
reward, fills the applications with data that form different states, and
implements the features that define the transitions. \emph{Reward
coverage} is how many goal-reaching trajectories the verifier rewards.
\emph{Data coverage} is how much application data the agent is exposed to
beyond the records the task reads or writes. \emph{Feature coverage} is
how many application features it implements beyond those the task uses.

\textbf{Existing works sacrifice each dimension to reduce construction cost.} On reward, some
choose to match actions step by step against a reference trajectory or to
train offline on annotated actions, which gives dense reward that is cheap
to
compute~\cite{lu2025uir1,im2025mobibench,zhang2024llamatouch}. On data,
some synthesize environments with a few generated records instead of
realistic application contents, which lowers cost and avoids copying real
user or copyrighted content: GUI-GENESIS fills a choice list with as few
as three to five plausible but incorrect
options~\cite{cao2026guigenesis}, and ScaleWoB starts from a minimal
environment with seed records~\cite{liu2026scalewob}. On features, a
builder can choose to implement only the functions that tasks use most,
which shortens construction. For example, GUI-GENESIS implements only the logic that
each task's interactions require~\cite{cao2026guigenesis}.

We answer this question in \gym{}~\cite{wu2026mobilegym}. Keeping task
instructions and native goal checks fixed, we adjust one dimension at a
time: single-path reward lowers reward coverage, record filtering
lowers data coverage, and functional pruning lowers feature coverage. We
train Qwen3-VL-4B-Instruct~\cite{bai2025qwen3vl} in each variant and
evaluate every policy in the original full environment (Full). We find that
lowering reward or data coverage trains a worse agent. Lowering feature coverage
trains a better one, and the gain comes from tasks whose environments
pruning left unchanged. A further experiment traces this gain to
how the objective weights episodes by length. Balancing every episode
reverses the gain, and balancing only the episodes on pruned tasks removes
it, whereas balancing only successes or only failures leaves it intact.
Without pruning, long successes that detour through unneeded functions
carry heavy weight, and the shared policy loses tasks it already solves to
loops. Pruning removes these detours.

These findings give a construction strategy: \textbf{a builder can omit
features the task does not need but should keep reward and data coverage
relatively complete.}
Following this strategy, we build \lite{}. It omits 87
template--module pairs across 72 of 192 training templates, removing
14.05\% of the application source lines per pruned task on average, while
keeping nearby navigation. Lite reaches \LiteSR\% success in the full
environment, 3.87 points above Full.

Our contributions are threefold. (1)~We propose the first
three-dimensional taxonomy of completeness in GUI environment
construction: reward coverage, data coverage, and feature coverage. We
measure how lowering each affects training and explain the result with a
mechanism experiment. (2)~We derive a construction strategy from these
findings as a methodology for building GUI environments. (3)~We build
\lite{} with this strategy, and it trains a stronger agent than the full
environment.

%% file: sections/02_related_work.tex
\section{Related Work}
\paragraph{GUI agents and their training.}
Executable benchmarks and platforms evaluate web, Android, and desktop
agents through
interaction~\cite{zhou2024webarena,koh2024visualwebarena,rawles2025androidworld,xie2024osworld,toyama2021androidenv,xu2025androidlab}.
Reinforcement learning in such environments trains device-control and GUI
agents~\cite{bai2024digirl,qi2025webrl,wang2025distrl,xu2026mobilerl,yang2025zerogui,wang2025uitars2}, and \gym{} provides
browser-hosted mobile applications with programmable state and parallel
rollouts for this purpose~\cite{wu2026mobilegym}. We use its editable
application logic to adjust each construction dimension during training
and restore the full environment for evaluation. On the objective side, Group Relative Policy
Optimization (GRPO) normalizes rewards within a
sampled group~\cite{shao2024deepseekmath}, and Group Sequence Policy
Optimization (GSPO) applies sequence-level importance ratios and
clipping~\cite{zheng2025gspo}. Prior work identifies token-length bias in
reasoning-model optimization~\cite{liu2025r1zero}. GUI interaction adds a
level of aggregation: one episode contains several model responses, each
with several tokens. We reweight responses per episode while keeping the
within-response objective, and test how this weighting changes the effect
of functional pruning.

\paragraph{Environment construction and reduction.}
GUI-GENESIS synthesizes task-specific environments with executable reward
assertions and compares them with real applications~\cite{cao2026guigenesis}, OS-Genesis derives tasks from exploration of existing
environments~\cite{sun2025osgenesis}, and ScaleWoB co-generates related tasks
to keep synthesized applications usable beyond one scripted
path~\cite{liu2026scalewob}. Other works replace real applications with
LLM-simulated, virtual, or mock
ones~\cite{wang2025uisimulator,wang2026stamp,tong2026maigym,xu2026mobileagentv35,tang2026phonebuddy}. On the evaluation side, LlamaTouch and
MobiBench show that single-path matching misjudges valid executions and
accept multiple paths through essential states or per-step
branches~\cite{zhang2024llamatouch,im2025mobibench}, recent verifiers
check goal-state anchors or probe the final
state~\cite{zhang2026gsar,cui2026agenticrm,wang2026vericua}, whereas offline RL still
rewards agreement with one annotated action~\cite{lu2025uir1}. Data scale
and device configuration also change how UI agents
generalize~\cite{li2024androidcontrol,lee2025bmoca}. These works
compare whole construction pipelines, evaluation rules, or data scale. We hold the applications and tasks fixed and
lower reward coverage, data coverage, and feature coverage one at a
time, so each outcome can be attributed to one dimension.

%% file: sections/03_study_design.tex
\section{Study Design}
\label{sec:design}
\subsection{Environment Model and Controlled Variants}
\paragraph{Environment model.}
We model a GUI environment as a POMDP
$(\mathcal{S},\mathcal{A},T,R,\Omega)$. A state $s\in\mathcal{S}$ holds the
application data and interface state. The policy receives only a
screenshot $o=\Omega(s)$ and emits a GUI action $a\in\mathcal{A}$, and $T$ is
the application's response. A task specifies an instruction $x$, an
initial state $s_0$, and a native goal check $V_x$, from which $R$ is
derived. A builder authors three parts per task, each defining one
dimension: the verifier defines $R$, the application data form $s_0$, and
the features define $T$. $\Omega$ renders the state and $\mathcal{A}$ is
fixed by the device.

\paragraph{Controlled variants.}
Each experiment group adjusts one dimension, keeps the other two as in
Full, and is evaluated with the original $R$, records, and functions. The
reduction is assigned per task template, a family of instructions and
initial states instantiated by a seed, so the rule stays fixed as task
instances vary. Each reduction is built not to reach into the other
dimensions. Separate manipulations
need not have separate effects: one policy learns all templates, so a
reduction on some tasks can change behavior on others, which the grouped
evaluation in \cref{sec:setup} measures.

\subsection{Coverage Dimensions and Their Adjustment}
\paragraph{Reward coverage.}
Reward coverage is how many goal-reaching trajectories, those that $T$
carries from $s_0$ to a state satisfying $V_x$, receive reward. To lower
it, we restrict reward to a reference operation sequence $c_x$:
\begin{equation}
 r_{\mathrm{path}}(\tau,x)
 =\mathbf{1}[V_x(\tau)=1]\,
  \mathbf{1}[M(\tau,c_x)=1],
 \label{eq:path}
\end{equation}
where $M$ matches ordered semantic operations, route, and target identity,
with limited motor fallbacks. One reference covers each training template.
Private DOM annotations identify operations for matching, while the policy
observes only screenshots (\appref{app:construction}).

The group keeps $s_0$ and $T$: all actions remain reachable and
divergence leaves the episode running, so every route stays executable
and only the set of rewarded paths shrinks. This is how reference-matching
procedures lower reward coverage in practice: a trajectory can satisfy
the goal and still receive zero reward.

\paragraph{Data coverage.}
Data coverage is how many \emph{records} the environment holds beyond
those the task reads or writes, such as unrelated alarms or messages.
Record filtering lowers it: R0, R1, and R2 retain protected records and
at most zero, one, or two unrelated records per reviewed list, changing 78
of the 192 training templates. Protected records cover both branches of a
conditional instruction, answer-relevant records, alternative entrances,
and shared helpers. Retained rows keep their order and identities.

The group keeps $R$ and $T$. Because it protects every record a solution
reads or writes, every route to the goal and every function on the
retained rows remain.

\paragraph{Feature coverage.}
Feature coverage is how many \emph{functions}, operations in $T$ whose
controls $\Omega$ renders, the environment implements beyond those the
task uses, such as the stopwatch. Functional pruning lowers it with
task-specific masks that disable dispensable function modules, such as
the stopwatch or Moments. \fzero{} disables all reviewed dispensable
modules for a template, while \fone{} retains $\lceil K/2\rceil$ of its
$K$ candidate modules by a stable hash. \fzero{} affects 78 templates,
with 127 template--module disablements across Clock, Notes, Calendar, and
WeChat. Masks keep a function whenever the task needs it: alarm tasks
lose the world-clock entrance while world-time tasks retain it.
Cross-application tasks combine profile rules and protect shared helpers, and replay and state checks verify the retained operations. Templates without
a reviewed rule use Full, and \appref{app:construction} lists all nine
profiles.

\subsection{Experimental Setup}
\label{sec:setup}
\paragraph{Training.}
We train Qwen3-VL-4B-Instruct on a dedicated \gym{} research split.
A pass comprises 16 outer steps of 12 tasks and eight rollouts per task,
with learning rate $10^{-6}$ and temperature 0.7.
Due to the page limit, we defer the remaining settings to \appref{app:setup}.

\paragraph{Evaluation.}
The 155 evaluation templates include 132 ID templates with new task
instances and 23 template-OOD templates. We train and evaluate under
seeds 42, 43, and 44, which yield 465 evaluation episodes per model, run
at temperature zero with a fixed simulator clock. Success requires goal
satisfaction and valid termination, and we also report goal satisfaction
alone. We report the mean over the three seeds at step~16, and
\cref{tab:seeds} in \appref{app:setup} gives its mean and standard
deviation across seeds.
Evaluating in Full tests whether a policy trained in a reduced variant
can operate when the original records and controls are present, as it
must when deployed on real applications, so any mismatch between training
and evaluation screens is part of the measured cost. Grouping evaluation
by whether a template's training environment was reduced shows whether a
gain comes from the modified tasks or from the others.

%% file: sections/04_findings.tex
\section{Which Construction Choices Hurt Performance?}
\label{sec:findings}
We lower each construction dimension in turn and evaluate the resulting
policy in the full environment. Lowering reward or data coverage lowers
success, whereas lowering feature coverage improves it.

\subsection{Narrow Reward Suppresses Valid Solutions}
\begin{table}[!htb]
 \centering
 \caption{\textbf{Lowering reward coverage.}
 Full-environment success over 465 instances.}
 \label{tab:path}
 \small
 \input{tables/path}
\end{table}

Reference-matching rewards are widely used to train GUI agents, offline
and online, because they are dense and hard to exploit
(\cref{sec:intro}). Single-path reward tests whether accepting one
anticipated solution costs a policy that explores on its own.

Under group-relative advantage, a rejected success can be pushed down.
At step~11 of alarm deletion, six rollouts delete the alarm through its
edit sheet and one uses long-press selection. All seven succeed in three
actions, and only the long-press path matches the reference. With one
failure in the group, each edit-sheet success receives advantage $-0.354$
and the reference-path success $+2.475$. Native goal reward on the same
samples gives all seven successes $+0.354$. Since the seven successes
share length and outcome, the sign reversal comes entirely from the
accepted path set. When every success in a group is rejected, the
restricted reward instead erases the success/failure contrast
(\appref{app:construction}).

The restricted policy finishes 10.53 points below Full: 191 versus 240
successes out of 465 (\cref{tab:path}). Narrowing the rewarded paths
therefore discards successful experience and penalizes valid alternatives,
even though every path remains executable. A training environment should
cover all feasible solution paths to provide the right learning signal.

\subsection{Fewer Records Lower Success}
Record filtering tests whether fewer unrelated records make learning
easier. Every filtering level instead ends below Full
(\cref{tab:resources}): at step~16, R0, R1, and R2 reach 47.96\%,
47.31\%, and 49.68\% against Full's 51.61\%, drops of 3.66, 4.30, and
1.93 points. At step~8 all three lead Full, with 45.59\%, 46.88\%, and 48.82\% against
43.44\%, but this early advantage reverses by the endpoint. The loss lies on the
132 templates seen in training, where R0, R1, and R2 solve 180, 182, and
188 of 396 instances against 198 for Full. On unseen templates they match
Full (38--43 against 42). On seen templates, the policy trained on
shortened lists meets the full lists at evaluation.

\begin{table}[t]
 \centering
 \caption{\textbf{Lowering data coverage.} Full-environment success (\%).
 R0/R1/R2 keep at most zero/one/two unrelated records per list. Seen/Unseen
 give step~16 successes on templates inside/outside the training set.}
 \label{tab:resources}
 \small
 \input{tables/resources}
\end{table}

Filtering records therefore costs where the filtered tasks meet real
record lists, and the environment we construct later keeps all
application records. \appref{app:history}
reports the intermediate checkpoints.

\subsection{Fewer Features Raise Success}
\subsubsection{Full-Environment Success}
Functional pruning tests whether removing dispensable functions costs
training. It does the opposite (\cref{tab:features}): F0 reaches 261/465 successes (\FzeroSR\%) against 240/465
(\FullSR\%) for Full, a 4.52-point gain. F1, which keeps roughly half of
the candidate functions, reaches 242/465.

\begin{table}[!htb]
 \centering
 \caption{\textbf{Lowering feature coverage.}
 Success (\%) over 465 full-environment instances.}
 \label{tab:features}
 \small
 \input{tables/features}
\end{table}

\subsubsection{Gains Lie in Unmodified Tasks}
\begin{figure}[t]
 \centering
 \includegraphics[width=\linewidth]{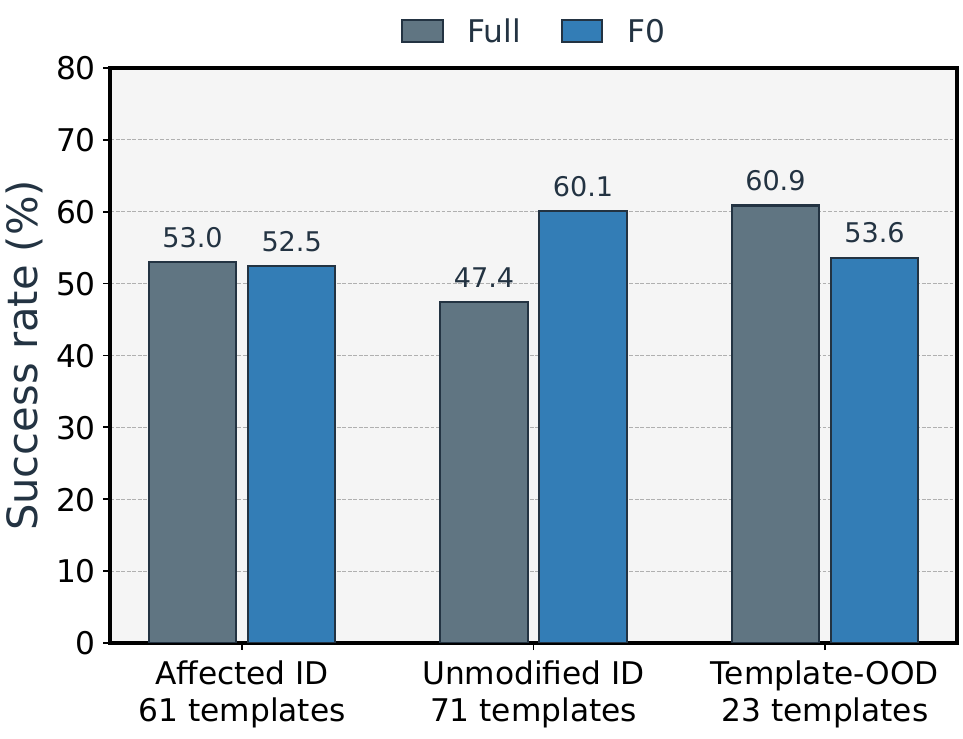}
 \caption{\textbf{Gains outside directly modified tasks.} Step~16 success
 rates (\%) in the full environment, grouped by the original F0 masks.}
 \label{fig:cohorts}
\end{figure}

The F0 gain lies entirely outside the tasks F0 modified. We partition
evaluation by the F0 masks (\cref{fig:cohorts}) into 61 affected
templates, whose training environments F0 pruned; 71 unmodified templates,
trained in identical environments under Full and F0; and 23 templates
unseen in training. Full and F0 solve 97 and 96 of 183 affected
instances, so pruning brings no net gain on the pruned tasks themselves. On the 213
unmodified instances, F0 raises successes from 101 to 128, while on unseen templates successes fall from 42 to 37. The net gain of 21 successes
thus comes from tasks whose environments F0 never changed. These
templates belong to applications outside the four pruned ones, so their
instructions and interfaces are identical in Full and F0: pruning
functions in some applications improves the agent on others.

The recovered successes often replace loops and step-limit failures:
F0 solves 46 instances on which Full loops or exhausts its action budget,
and introduces 27 such failures. Joint successes have similar lengths. \Cref{sec:mechanism} tests how training on the pruned tasks
affects capabilities learned on the others.

%% file: tables/path.tex
\begin{tabular}{@{}lrr@{}}
\toprule
Training & Success (\%) & Successes \\
\midrule
Base & 38.71 & 180 \\
Full & 51.61 & 240 \\
Path-restricted & 41.08 & 191 \\
\bottomrule
\end{tabular}

%% file: tables/resources.tex
\begin{tabular}{@{}lrrrr@{}}
\toprule
Condition & Step8 & Step16 & Seen /396 & Unseen /69 \\
\midrule
Full & 43.44 & 51.61 & 198 & 42 \\
R0 & 45.59 & 47.96 & 180 & 43 \\
R1 & 46.88 & 47.31 & 182 & 38 \\
R2 & 48.82 & 49.68 & 188 & 43 \\
\bottomrule
\end{tabular}

%% file: tables/features.tex
\begin{tabular}{@{}lrrr@{}}
\toprule
Environment & Step8 & Step12 & Step16 \\
\midrule
Full & 43.44 & 46.45 & 51.61 \\
F0 & 44.95 & 48.17 & 56.13 \\
F1 & 48.17 & 41.29 & 52.04 \\
\bottomrule
\end{tabular}

%% file: sections/06_mechanism.tex
\begin{figure*}[t]
 \centering
 \includegraphics[width=\textwidth]{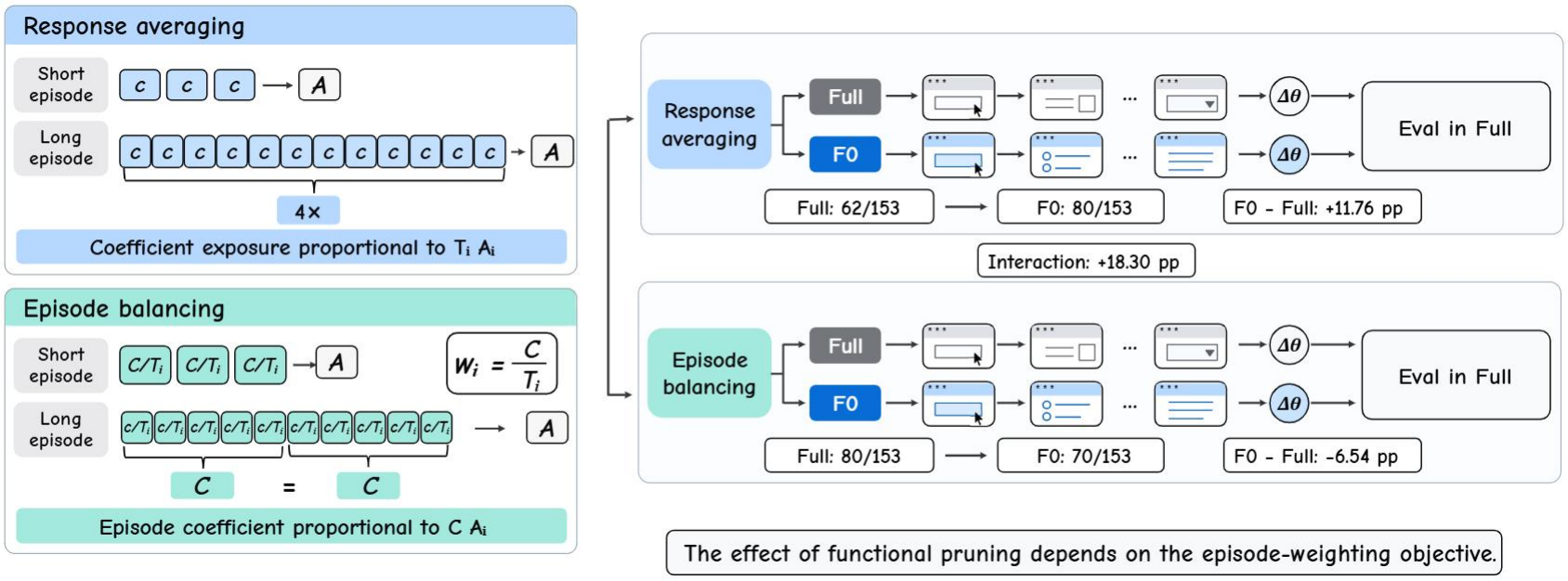}
 \caption{\textbf{Episode weighting changes the environment effect.}
 Left: response averaging gives longer episodes more advantage-bearing terms, and episode balancing removes this dependence with weights $C/T_i$.
 Right: four continuations start from the same F0 step~12 model. Counts show successes
 on 153 retention instances evaluated in Full. Weighting reverses the F0--Full
 gap, yielding an 18.30-point interaction.}
 \label{fig:mechanism}
\end{figure*}

\section{Why Fewer Features Help}
\label{sec:mechanism}
\subsection{Episode Length and Update Weight}
The F0 gain on unchanged tasks implies that pruning acts through the shared
policy's updates. Pruning changes training trajectories even with a fixed
task set and terminal reward. Episode $i$ contains $T_i$ valid GUI responses and
receives a terminal group-relative advantage $A_i$, copied to each response.
Each response can contain multiple text tokens. Under response averaging,
the episode's raw coefficient exposure is proportional to $T_i A_i$.

Twelve responses repeat an advantage four times as often as three, for
either sign (\cref{fig:mechanism}, left). By changing the screens and
actions the policy visits, function masks can change how much each episode
contributes to training.

We test this route by weighting each valid response in episode $i$ by
\begin{equation}
 w_i=\frac{C}{T_i},\qquad C=\frac{1}{N}\sum_{j=1}^{N}T_j.
 \label{eq:weight}
\end{equation}
Here $N$ counts episodes with at least one valid response. For the
token-averaged response objective $\ell_{it}$, this changes aggregation from
\begin{align}
 \frac{1}{\sum_i T_i}\sum_i\sum_{t=1}^{T_i}\ell_{it}
 \quad\text{to}\quad
 \frac{1}{N}\sum_i\frac{1}{T_i}\sum_{t=1}^{T_i}\ell_{it}.
 \label{eq:objectives}
\end{align}
We call the latter \emph{episode balancing}. It rescales advantages after
group normalization and leaves the mean weight over valid responses at one.
Each episode's coefficient becomes $CA_i$, so response count no longer
multiplies its advantage. Positive weights preserve advantage signs and
scale the clipped response objective by the same factor. The implementation
excludes padding and leaves within-response token averaging unchanged.
\appref{app:weighting} derives the identity and describes the implemented
loss and realized training budgets.

\subsection{Crossing Environment and Weighting}
We cross the training environment (Full or F0) with episode
weighting (response averaging or episode balancing).
Each arm restores the same F0 step~12 model and optimizer state and processes
steps~13--16: 19 function-pruned templates and 29 other templates.
The retention set contains 51 originally unmodified templates practiced
by step~12 and absent from every tail. Evaluating the step-12 start and the
four endpoints measures what survives further updates on other tasks.

Each continuation samples its own trajectories on the same remaining task sequence.
Comparing F0 with Full within each weighting condition gives
two environment gaps, and their difference measures the interaction. Within an
environment, the two arms share start state, task sequence, and optimizer settings,
and differ only in whether $T_i$ multiplies each advantage, so the
interaction is the effect of length weighting on the environment gap.
Omitting the retention templates from every tail makes this a test of
capability under updates on other tasks. A final success can reflect either
a preserved solution or a newly acquired one, so we also compare each
instance with its outcome at the common start. These paired transitions
separate preserved capability from indirectly acquired successes.

\begin{table}[t]
 \centering
 \caption{\textbf{Training from the F0 step~12 model.}
 Ret./Goal count success/native goal satisfaction on 153 retention instances. Tail environment and weighting
 (response averaging or episode balancing) vary after step~12.}
 \label{tab:mechanism}
 \small
 \input{tables/mechanism}
\end{table}

Episode weighting changes the effect of the training environment
(\cref{tab:mechanism,fig:mechanism}). On the 153 primary instances, the
F0-minus-Full success gap is +11.76 points under response averaging and
$-6.54$ points under episode balancing. Their difference is +18.30 points:
\begin{equation}
 \frac{(80-62)-(70-80)}{153}\times100\approx18.30.
 \label{eq:interaction}
\end{equation}
The interaction is positive at all three seeds. Goal satisfaction
gives a similar 18.95-point interaction. Thus the reversal also appears
when we score the final task state separately from valid termination.

\subsection{Capability Retention}
\paragraph{Retained and new successes.}
Under response averaging, F0 nearly preserves the target success count
through late training. The step-12 start succeeds on 81/153 targets; F0
ends with 80 and Full with 62. Full loses 24 initial successes and gains
five, whereas F0 loses 15 and gains 14, so F0's 18-success lead splits
into nine more preserved and nine more newly solved instances. Under
episode balancing, Full ends with 80 and F0 with 70, and Full's
ten-success lead splits into eight more preserved and two more newly
solved instances. The environment ordering therefore reverses in
preserved capability as well as in the combined endpoint count
(\appref{app:weighting}).

\paragraph{Advantage sign of long episodes.}
F0 does not help by producing shorter episodes. The response-averaged Full
and F0 tails solve 142 and 141 of 384 training episodes, and F0 generates
more GUI responses (4196 versus 3997). What differs is the sign that long
episodes carry. On affected templates, we sum $T_iA_i$ over positive
advantages and $T_i|A_i|$ over negative ones: positive exposure is
386.2 for Full and 320.3 for F0, negative exposure 416.3 and 500.8.

Pruning therefore changes which episodes become long on the pruned tasks.
In Full, more of the length-weighted mass sits in successes that reach the
goal after detours through functions the task does not need, so response
averaging reinforces long, wandering action sequences. With those functions
removed, F0 has fewer such successes, and its long episodes are mostly
failures. The same policy serves every application, so reinforced detours
can reach tasks it already solves: in Full, 21 of the 24 lost successes
become loops.

\begin{table}[t]
 \centering
 \caption{\textbf{Sign-specific episode balancing.} Successes on the 153
 retention instances. The sign rows apply $w_i=C/T_i$ only to episodes
 with the given advantage sign and keep response averaging for the rest. The
 last two rows balance only episodes from the 19 affected or the 29 other
 templates, with $C$ averaged over those episodes, and response-average the
 rest.}
 \label{tab:signweighting}
 \small
 \input{tables/sign_weighting}
\end{table}

\paragraph{Source of Full's loss.}
Two partial forms of balancing locate this effect
(\cref{tab:signweighting}). Balancing one advantage sign leaves the gap
nearly intact: balancing negative episodes moves Full from 62 to 66 and
leaves F0 at 80, and balancing positive episodes gives 64 and 81.
Balancing both signs on the 19 affected templates alone, with $C$ averaged
over their episodes and response averaging on the other 29, closes the gap
to +1. Full rises from 62 to 79, recovering 17 of the 18 successes that
full balancing restores, while F0 stays at 80. The reverse split, which
balances the 29 other templates and response-averages the affected ones,
leaves Full at 64 and lowers F0 to 71, a +7 gap. \emph{Full's retention
loss thus comes from length weighting on the pruned tasks, and removing it
requires balancing both signs there.} Response-averaged F0 does not depend
on this weighting: pruning helps by removing the episodes whose length
weighting causes Full's loss, not by adding a training signal of its own.
F0's drop under full balancing comes from the other templates instead, so
it does not bear on the F0 gain.

\paragraph{Examples and other explanations.}
Paired trajectories illustrate how the interaction appears in behavior.
For a Railway ticket-price query, response-averaged F0 and balanced Full
retrieve an order and answer in three actions, while the other two arms loop on
date selection. A Reading privacy-setting task shows the opposite pattern:
response-averaged F0 reaches the step limit while the other arms succeed.
\appref{app:cases} provides both examples and their selection rule.

\appref{app:alternatives} rules out two other explanations: F0 policies do
not gain from being evaluated on the simpler F0 screens, and F0 does not
lead before the unmodified tasks are practiced. \emph{Functional
pruning therefore helps by preserving previously learned capabilities, and
this benefit depends on how episodes on the pruned tasks are weighted by
length.}

%% file: tables/mechanism.tex
\begin{tabular}{@{}llrr@{}}
\toprule
Tail env. & Weighting & Ret. & Goal \\
\midrule
Step-12 start & -- & 81 & 83 \\
Full & Response & 62 & 65 \\
F0 & Response & 80 & 82 \\
Full & Episode & 80 & 84 \\
F0 & Episode & 70 & 72 \\
\bottomrule
\end{tabular}

%% file: tables/sign_weighting.tex
\begin{tabular}{@{}lrrr@{}}
\toprule
Weighting & Full & F0 & F0$-$Full \\
\midrule
Response averaging & 62 & 80 & +18 \\
Episode balancing & 80 & 70 & $-$10 \\
Balance $A_i<0$ only & 66 & 80 & +14 \\
Balance $A_i>0$ only & 64 & 81 & +17 \\
Balance affected only & 79 & 80 & +1 \\
Balance other only & 64 & 71 & +7 \\
\bottomrule
\end{tabular}

%% file: sections/05_lite.tex
\begin{figure*}[t]
 \centering
 \includegraphics[width=\textwidth]{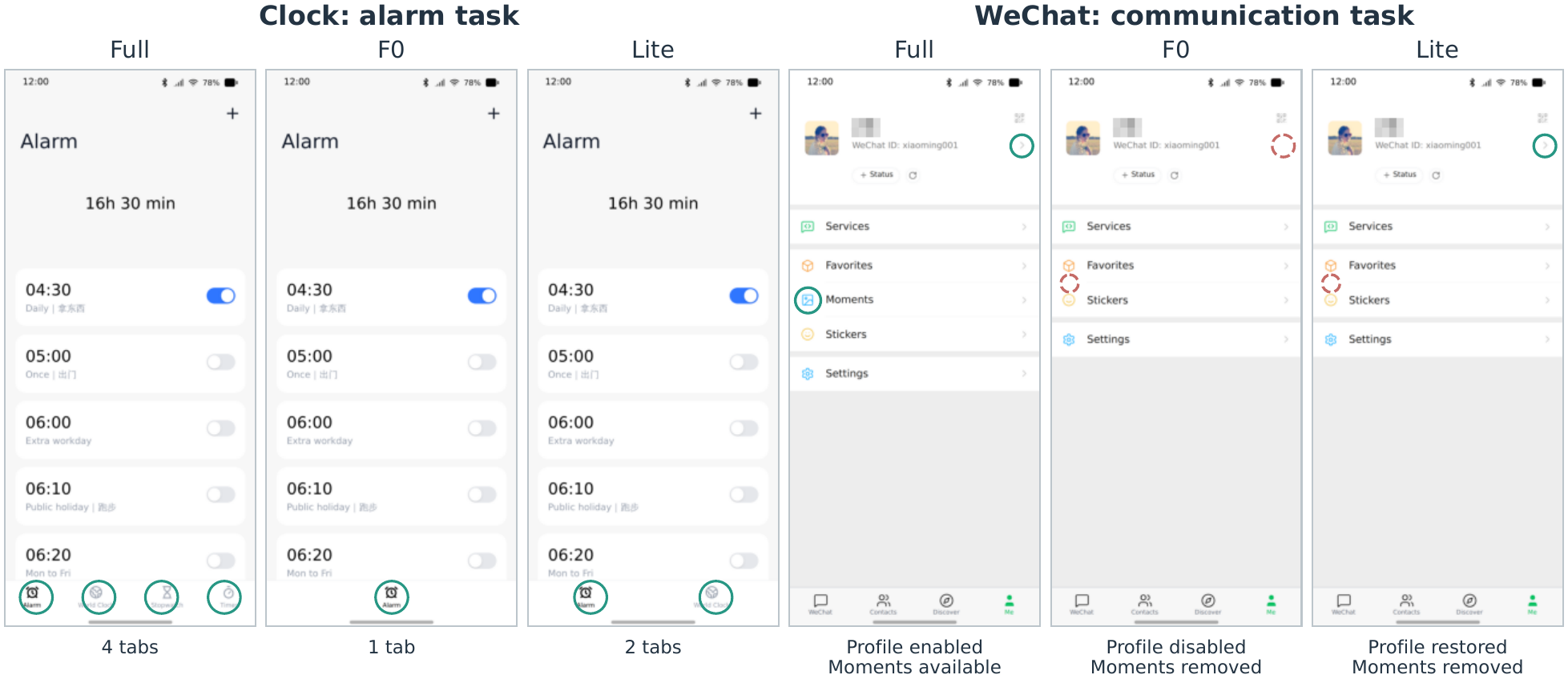}
 \caption{\textbf{Selective functional pruning in MobileGym-Lite.}
 Full-screen screenshots use identical task states. Clock retains four tabs
 in Full, one in F0, and two in Lite. WeChat retains the profile card:
 F0 disables clicking and removes its chevron, and Lite restores both.
 Moments is removed in F0 and Lite. Circles highlight these controls, and dashed circles mark missing WeChat entrances.}
 \label{fig:screens}
\end{figure*}

\section{Construction Strategy and MobileGym-Lite}
\label{sec:lite}
\subsection{Construction Strategy}
The study gives one rule per dimension.
(1)~\textbf{Keep reward coverage complete:} the verifier rewards any
trajectory that passes the native goal check.
(2)~\textbf{Keep data coverage complete:} the environment retains every
application record.
(3)~\textbf{Lower feature coverage selectively:} a builder prunes
dispensable functions and evaluates the pruned environment together with
its training objective, since \cref{sec:mechanism} shows that the objective
decides whether pruning preserves learned capabilities.

\subsection{MobileGym-Lite}
\paragraph{Selective pruning.}
Following this strategy, \lite{} rewards any trajectory that passes the native
goal check, retains every application record, and prunes dispensable
functions.

It removes dispensable functions but keeps
nearby navigation and shared operations. Starting from F0, it restores
world-clock access for alarm tasks, Calendar settings for date-tool tasks,
and profile editing for WeChat communication tasks. Alarm tasks thus keep
the alarm and world-clock tabs but lose stopwatch and timer entrances, and communication tasks keep profile editing but lose Moments. These
restorations affect 36 templates, six of which then retain all functions.
\appref{app:construction} specifies all profiles and restorations.

\paragraph{Savings.}
Lite omits 87 template--module pairs across 72 of the 192 training
templates: for these tasks, the corresponding functions need not be
provided. Measured in lines of code, the application source a pruned task
needs shrinks by 14.05\% on average over these 72 templates. Building in
this style also costs less: we had Codex rebuild ten \gym{}
applications, including WeChat and Bilibili, once completely in \gym{}'s
style and once implementing only the functions the main task templates
use, close to F0, and the second rebuild used about 33\% fewer tokens.
On nine fixed screens, visible control nodes drop from 126 to
117, a 7.14\% reduction (F0: 113), while keeping the world-clock tab and
the WeChat profile entrance (\cref{fig:screens}). Lite keeps all 192
training templates and all application records. \appref{app:screens} gives the per-screen measurements.

\subsection{Late-Training Stabilization}
\label{sec:clipping}
\paragraph{Instability under reduction.}
Every reduced environment we trained, whether it lowers reward, data, or
feature coverage, shows a surge of loops and format errors late in
training. From step~8 to step~16, loop terminations rise from 111 to 135
under R0 and format errors from 12 to 175 under R1, while Full's loop
terminations fall from 117 to 109. The path-restricted policy ends with
144 loop terminations against 109 for its Full reference. Under feature
pruning, success also fluctuates between checkpoints: F1 falls from 224
to 192 between steps~8 and 12 before recovering to 242, and Lite from 239
to 233 before recovering to 250.

\paragraph{Narrow late clipping.}
We reduce this with one adjustment: after step~$k$, the GSPO
importance-ratio clip offsets narrow from the default 0.2/0.28
(lower/upper, W) to 0.002/0.0028 (N), written
$\mathrm{W}{\to}\mathrm{N}_{k}$. Branches share model and optimizer state
at step~8 for Full, F0, and Lite (\cref{tab:clipping}). The adjustment
smooths Lite's trajectory, from 239, 233, and 250 to 239, 242, and 258
at steps~8, 12, and 16, and raises its endpoint by eight successes. It
leaves F0 at 262 against 261, and brings Full, which trains stably, no
gain (238 against 240). Lite therefore uses $\mathrm{W}{\to}\mathrm{N}_{8}$,
and all other main-text runs use W throughout.

\begin{table}[t]
 \centering
 \caption{\textbf{Narrow late clipping.} Both branches of a row start
 from the same model and optimizer state at the split step. W/N give step~16 successes
 under wide/narrow clipping thereafter, and $\Delta$ is N minus W in points.}
 \label{tab:clipping}
 \small
 \setlength{\tabcolsep}{4pt}
 \input{tables/clipping}
\end{table}

\subsection{Full-Environment Performance}
\paragraph{Success.}
Pruning speeds up learning (\cref{tab:lite}). Full and Lite train under
the same settings through step~8, and Lite already solves 239 of 465
instances against Full's 202, close to Full's step-16 result of 240.
At step~16, with $\mathrm{W}{\to}\mathrm{N}_{8}$, Lite reaches 258/465
successes (\LiteSR\%), 18 more than Full (\FullSR\%).

\begin{table}[t]
 \centering
 \caption{\textbf{MobileGym-Lite in the full environment.}
 Success (\%) over 465 instances. Both runs share settings through
 step~8; Lite narrows the importance-ratio clip thereafter
 (\cref{sec:clipping}).}
 \label{tab:lite}
 \small
 \input{tables/lite}
\end{table}

\paragraph{Where the gain lies.}
Lite gains in all three evaluation groups: one success on the 183
affected instances, 12 on the 213 unmodified ones, and five on the 69
unseen-template ones. It solves 35 instances on which Full loops or
exhausts the action budget, while introducing 14 such failures.

\paragraph{Why Lite.}
F0 reaches 261 successes, three more than Lite. We release Lite because it
prunes more conservatively: it restores three modules next to the tasks'
own navigation and disables 87 rather than 127 template--module pairs, and
its gain holds in every evaluation group, whereas F0 loses five successes
on unseen templates and one on affected ones. Lite instantiates the construction rule rather than
the highest-scoring mask.

%% file: tables/clipping.tex
\begin{tabular}{@{}lrrrr@{}}
\toprule
Environment & Split step & W /465 & N /465 & $\Delta$ pp \\
\midrule
Full & 8 & 240 & 238 & -0.43 \\
F0 & 8 & 261 & 262 & +0.22 \\
Lite & 8 & 250 & 258 & +1.72 \\
\bottomrule
\end{tabular}

%% file: tables/lite.tex
\begin{tabular}{@{}lrrr@{}}
\toprule
Environment & Step8 & Step12 & Step16 \\
\midrule
Full & 43.44 & 46.45 & 51.61 \\
Lite & 51.40 & 52.04 & 55.48 \\
\bottomrule
\end{tabular}

%% file: sections/07_discussion.tex
\section{Discussion and potential impact}
This study turns environment construction from engineering trade-offs
into measured decisions, because the dimensions differ in what a
reduction removes. Leaving valid
solutions unrewarded removes training signal and penalizes valid
alternatives, and filtered records cost success where they return at
deployment, whereas unneeded features mainly add detours. A builder
should therefore keep reward and data coverage complete and can lower
feature coverage, spending the saved effort on more applications and
tasks.

The mechanism adds two criteria for judging a reduced environment. Its
gain appears on unmodified tasks, so a multi-task environment should be
judged by what the agent retains on other tasks, not only by what it
learns on the reduced ones. The gain also depends on episode weighting,
so a pruned environment should be judged together with its training
objective.

Our reward-coverage study narrows the rewarded trajectories but keeps the
transitions $T$; removing alternative routes from $T$ while keeping every
task solvable is left open. We use one base model and one simulator, and
replication across starting policies and simulators is the next step.

%% file: sections/08_conclusion.tex
\section{Conclusion}
A less complete GUI environment need not train a worse agent. What matters is which dimension a construction choice lowers. Lowering reward
coverage suppresses valid solutions, lowering data coverage lowers success
on the filtered tasks, and lowering feature coverage improves
full-environment success,
entirely on tasks it never modified. The gain comes from length weighting
on the pruned tasks: without pruning, long detour successes there are
reinforced and erode tasks the policy already solves, and balancing only
those tasks removes the gap. \lite{} applies these
findings and reaches \LiteSR\% success in the full environment, above
Full's \FullSR\%.

%% file: appendix.tex
\clearpage
\appendix
\setcounter{page}{1}
\renewcommand{\thepage}{S\arabic{page}}
\setcounter{figure}{0}
\setcounter{table}{0}
\setcounter{equation}{0}
\renewcommand{\thefigure}{S\arabic{figure}}
\renewcommand{\thetable}{S\arabic{table}}
\renewcommand{\theequation}{S\arabic{equation}}
\renewcommand{\theHfigure}{supp.\arabic{figure}}
\renewcommand{\theHtable}{supp.\arabic{table}}
\renewcommand{\theHequation}{supp.\arabic{equation}}
\ifdefined\standalonesupplement
  \maketitle
\else
  \maketitlesupplementary
\fi
\input{sections/09_supplement}

%% file: sections/09_supplement.tex
This supplement provides task and evaluation definitions (\cref{app:setup}),
construction rules (\cref{app:construction}), complete results
(\cref{app:history}), interface measurements (\cref{app:screens}),
episode-weighting analysis (\cref{app:weighting}), alternative mechanism
checks (\cref{app:alternatives}), and paired examples (\cref{app:cases}).

\input{sections/appendix/a_setup}
\input{sections/appendix/b_construction}
\input{sections/appendix/c_results}
\input{sections/appendix/d_screens}
\input{sections/appendix/e_weighting}
\input{sections/appendix/f_alternatives}
\input{sections/appendix/g_cases}

%% file: sections/appendix/a_setup.tex
\section{Tasks and Evaluation}
\label{app:setup}
The study uses 192 training templates and 155 evaluation templates.
The overlap contains 132 templates, evaluated with newly sampled instances, and the remaining 23 evaluation templates form the template-OOD set.
Each template specifies an instruction, initial state, action budget, and
executable goal checks.

Environment labels identify available records/functions, and training branches also specify clipping and starting state. A continuation restores a saved
checkpoint. We train and evaluate under seeds 42, 43, and 44.
\Cref{tab:seeds} gives the mean and standard deviation of step-16
success across seeds for Full, Lite, and F0.

\begin{table}[t]
 \centering
 \caption{\textbf{Step-16 success across seeds.} Mean $\pm$ standard
 deviation over seeds 42, 43, and 44, in points. Clipping follows
 \cref{tab:history}.}
 \label{tab:seeds}
 \small
 \input{tables/seed_spread}
\end{table}

\begin{table}[b]
 \centering
 \caption{\textbf{Shared settings for the functional-pruning and mechanism studies.}
 An outer step samples 12 task instances, each with eight rollouts.}
 \label{tab:settings}
 \small
 \begin{tabular}{@{}ll@{}}
 \toprule
 Setting & Value \\
 \midrule
 Base model & Qwen3-VL-4B-Instruct \\
 Training templates / outer steps & 192 / 16 \\
 Tasks per step / rollouts per task & 12 / 8 \\
 Learning rate & $10^{-6}$ \\
 Training temperature / token cap & 0.7 / 1024 \\
 Evaluation temperature / token cap & 0 / 2048 \\
 Evaluation top-$p$ & 0.95 \\
 Seeds & 42, 43, 44 \\
 Simulator time (Unix ms) & 1789099200000 \\
 \bottomrule
 \end{tabular}
\end{table}

Task membership is fixed across the functional-pruning comparisons.
The original F0 masks partition the evaluation set into 61 affected ID,
71 unmodified ID, and 23 template-OOD templates, giving 183, 213, and 69
instances. The mechanism study's primary retention set consists of 51 of
the 71 unmodified ID templates. These have all received practice by the
common step~12 checkpoint and receive none in steps~13--16.
The remaining curriculum contains 48 templates: 19 affected and 29 others.

Success requires native goal satisfaction and valid episode termination. We also report native-goal success separately in the mechanism study.
Full-environment evaluation uses three seeds: 465 episodes per model and
153 on the retention set.

%% file: tables/seed_spread.tex
\begin{tabular}{@{}llr@{}}
\toprule
Environment & Clip & Success (\%) \\
\midrule
Full & W & $51.61\pm1.02$ \\
Lite & $\mathrm{W}{\to}\mathrm{N}_{8}$ & $55.48\pm0.96$ \\
F0 & $\mathrm{W}{\to}\mathrm{N}_{8}$ & $56.34\pm0.99$ \\
\bottomrule
\end{tabular}

%% file: sections/appendix/b_construction.tex
\section{Environment Construction}
\label{app:construction}
\subsection{Single-Path Reward}
A reference is an ordered sequence of semantic operations tied to a task
instance. The bundle covers all 192 training templates:
139 references are semantic, 46 combine semantic operations with motor
fallbacks, and seven use the historical motor representation.
The 1546 retained operations include 181 motor fallbacks. Matching preserves
operation order, route, and target identity. Equivalent pure-scroll chunks
can be merged after replay, while unresolved motor targets use narrow
observed regions. Private matching annotations are separate from the
screenshots supplied to the policy.

The learning effect of a rejected success depends on its sampled group.
For a driving-cost query, six successful CLICK-then-TYPE solutions differ
from the reference's direct TYPE, so all eight restricted rewards become zero,
erasing the group's success/failure contrast.
Birthday creation exposes sensitivity to an extra focus click and the
save-button context. In a Weather group, all eight valid solutions are
rejected, but native binary success already gives zero relative advantage
because all eight rewards agree. These cases separate reversal of useful
credit, removal of reward contrast, and rejection without a changed contrast.

\subsection{Record Filters}
Record filtering operates on reviewed lists of application records.
A protected set contains records needed by either branch of a conditional
task, its answer, alternate entrances, or shared helpers. R0/R1/R2 retain
that protected set and up to 0/1/2 unrelated records per list. Retained rows
preserve their complete fields, identities, and order. Filtering
changes 78 training templates and leaves other templates unchanged.

\subsection{Function Profiles and Lite Restorations}
Function masks are unions of reviewed profile rules
(\cref{tab:profiles}). A cross-application template may use more than one
profile. F0 disables every candidate module in that union. F1 ranks the
candidates by a stable hash of the version, template identifier, and module
identifier, then keeps $\lceil K/2\rceil$ of the $K$ candidates.
Lite starts from the F0 union and restores the modules in the table's last
column. Tasks outside these profiles keep all functions.

\begin{table*}[t]
 \centering
 \caption{\textbf{Complete function profiles and selective restorations.}
 F0 disables the listed functions for templates assigned to each profile.
 Lite restores only the functions in the final column. Profile rules are
 combined for tasks that use multiple applications.}
 \label{tab:profiles}
 \small
 \begin{tabular}{@{}lll@{}}
 \toprule
 Task profile & Functions disabled by F0 & Restored in Lite \\
 \midrule
 Clock: alarm & World clock, stopwatch, timer & World clock \\
 Clock: world time & Stopwatch, timer & -- \\
 Notes: note operations & Todo list & -- \\
 Notes: todo operations & Notes settings & -- \\
 Calendar: fixed events/settings & Date calculator & -- \\
 Calendar: date tools & Calendar settings & Calendar settings \\
 WeChat: communication & Moments, profile editing & Profile editing \\
 WeChat: personal settings & Moments & -- \\
 WeChat: social functions & Profile editing & -- \\
 \bottomrule
 \end{tabular}
\end{table*}

Runtime gates consume module identifiers, leaving task instructions,
parameters, records, and native goal checks unchanged. Masks reset between
episodes. The three Lite restorations change 36 templates relative to F0, and six of these then have no remaining disablements. Overall, F0 has 127
template--module disablements across 78 templates, and Lite has 87 across
72. These profile-level choices preserve the task set while changing the
functionality encountered during interaction.

%% file: sections/appendix/c_results.tex
\section{Complete Results and Training Curves}
\label{app:history}
\subsection{Environments, Schedules, and Branch Origins}
\begin{figure}[t]
 \centering
 \includegraphics[width=\linewidth]{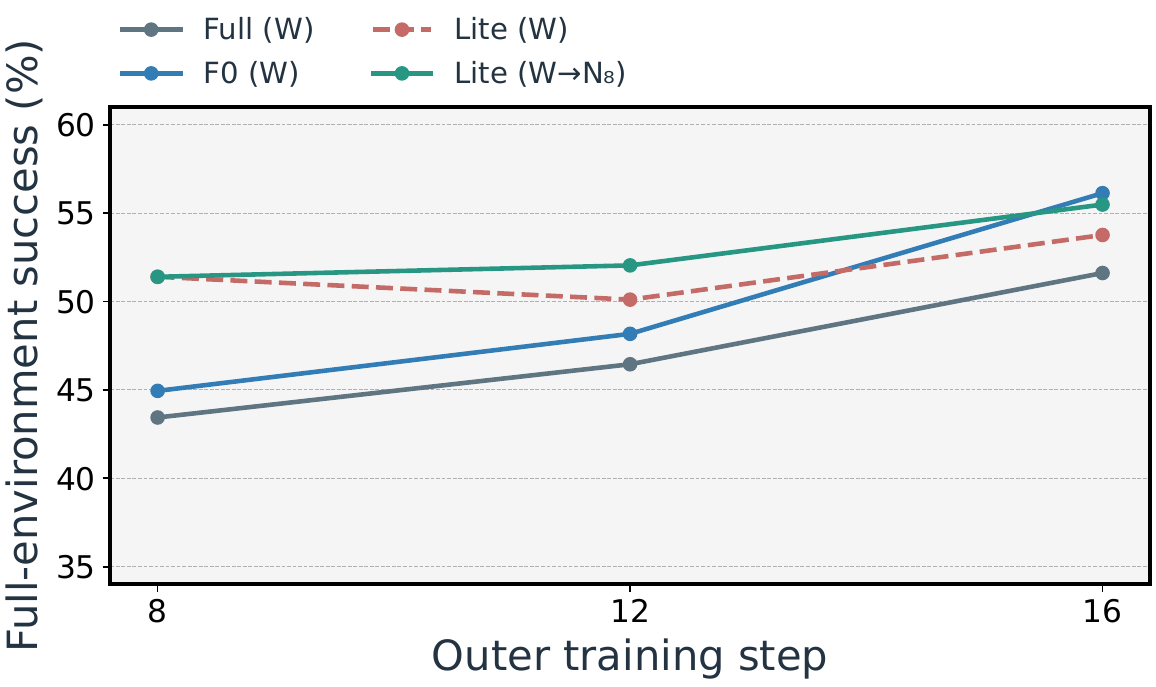}
 \caption{\textbf{Environment and clipping schedule through training.}
 Full-environment success. W denotes wide clipping and
 $\mathrm{W}{\to}\mathrm{N}_{8}$ narrow clipping after step~8.
 Both Lite curves share the same step~8 checkpoint.}
 \label{fig:curves}
\end{figure}

\Cref{tab:history} lists all 7 evaluated training runs. Each row fixes
an environment, a clipping schedule, and a starting state. The environments
differ only in which function modules their masks disable. Records,
instructions, and goal checks are those of Full.
\begin{itemize}
 \item \textbf{Full} disables nothing.
 \item \textbf{F0} disables every reviewed dispensable module
  (78 templates, 127 disablements). \textbf{F1} keeps
  $\lceil K/2\rceil$ of each template's $K$ candidate modules.
 \item \textbf{Lite} is F0 with three modules restored (world clock,
  Calendar settings, WeChat profile editing): 72 templates, 87
  disablements.
\end{itemize}
Full~(W) and
F0~(W) are the reference runs of Section~4. Rows with a branch start other
than the base model continue from the named checkpoint. Section~6.3
defines the clipping schedules, and \cref{fig:curves} plots success
through training under each.

\begin{table*}[t]
 \centering
 \caption{\textbf{Complete functional-pruning results with branch origins.}
 Step columns give success counts out of 465, and ID/OOD columns describe step~16.
 W means wide clipping throughout, and $\mathrm{W}{\to}\mathrm{N}_{k}$ switches after step~$k$.
 Superscript $s$ marks the shared source score and a dash an unevaluated checkpoint.}
 \label{tab:history}
 \small
 \setlength{\tabcolsep}{4pt}
 \input{tables/history}
\end{table*}

\subsection{Template Groups and F0 Summary}
The fixed-group breakdown localizes the endpoint improvements
(\cref{tab:cohortcounts}). Relative to Full, F0 gains 27 successes on
unmodified tasks, loses one on affected tasks, and loses five on
unseen templates. Lite gains 12, one, and five, respectively.

\begin{table}[t]
 \centering
 \caption{\textbf{Endpoint success by original F0 groups.}
 Counts are out of 183 affected, 213 unmodified, and 69 unseen-template instances, and All sums the three groups. Clipping follows
 \cref{tab:history}.}
 \label{tab:cohortcounts}
 \small
 \setlength{\tabcolsep}{4pt}
 \input{tables/cohort_counts}
\end{table}

\subsection{Data-Coverage Results}
R0, R1, and R2 reach 223, 220, and 231 successes at step~16, against 240
for Full.

\subsection{Reward-Coverage Results}
The path-restricted run contains 1536 training episodes: 600 native
successes, 348 reference matches, and 333 episodes satisfying both conditions.
Outer steps~3 and 16 have zero group-relative advantage.
Final evaluation gives 191 successes out of 465. The within-group cases in
\cref{app:construction} characterize how the single-path reward changes
credit assignment.

%% file: tables/history.tex
\begin{tabular}{@{}lllrrrrr@{}}
\toprule
Environment & Clip & Branch start & Step8 & Step12 & Step16 & ID /396 & OOD /69 \\
\midrule
Full & W & Base model & 202 & 216 & 240 & 198 & 42 \\
F0 & W & Base model & 209 & 224 & 261 & 224 & 37 \\
F0 & $\mathrm{W}{\to}\mathrm{N}_{8}$ & F0 (W): step8 & 209 \textsuperscript{s} & -- & 262 & -- & -- \\
F1 & W & Base model & 224 & 192 & 242 & 201 & 41 \\
Full & $\mathrm{W}{\to}\mathrm{N}_{8}$ & Full: step8 & 202 \textsuperscript{s} & 213 & 238 & 193 & 45 \\
Lite & W & Base model & 239 & 233 & 250 & 206 & 44 \\
Lite & $\mathrm{W}{\to}\mathrm{N}_{8}$ & Lite (W): step8 & 239 \textsuperscript{s} & 242 & 258 & 211 & 47 \\
\bottomrule
\end{tabular}

%% file: tables/cohort_counts.tex
\begin{tabular}{@{}llrrrr@{}}
\toprule
Environment & Clip & Aff. & Unmod. & OOD & All \\
\midrule
Full & W & 97 & 101 & 42 & 240 \\
F0 & W & 96 & 128 & 37 & 261 \\
Lite & $\mathrm{W}{\to}\mathrm{N}_{8}$ & 98 & 113 & 47 & 258 \\
\bottomrule
\end{tabular}

%% file: sections/appendix/d_screens.tex
\section{Pruned Interfaces}
\label{app:screens}
\begin{figure*}[t]
 \centering
 \includegraphics[width=\textwidth]{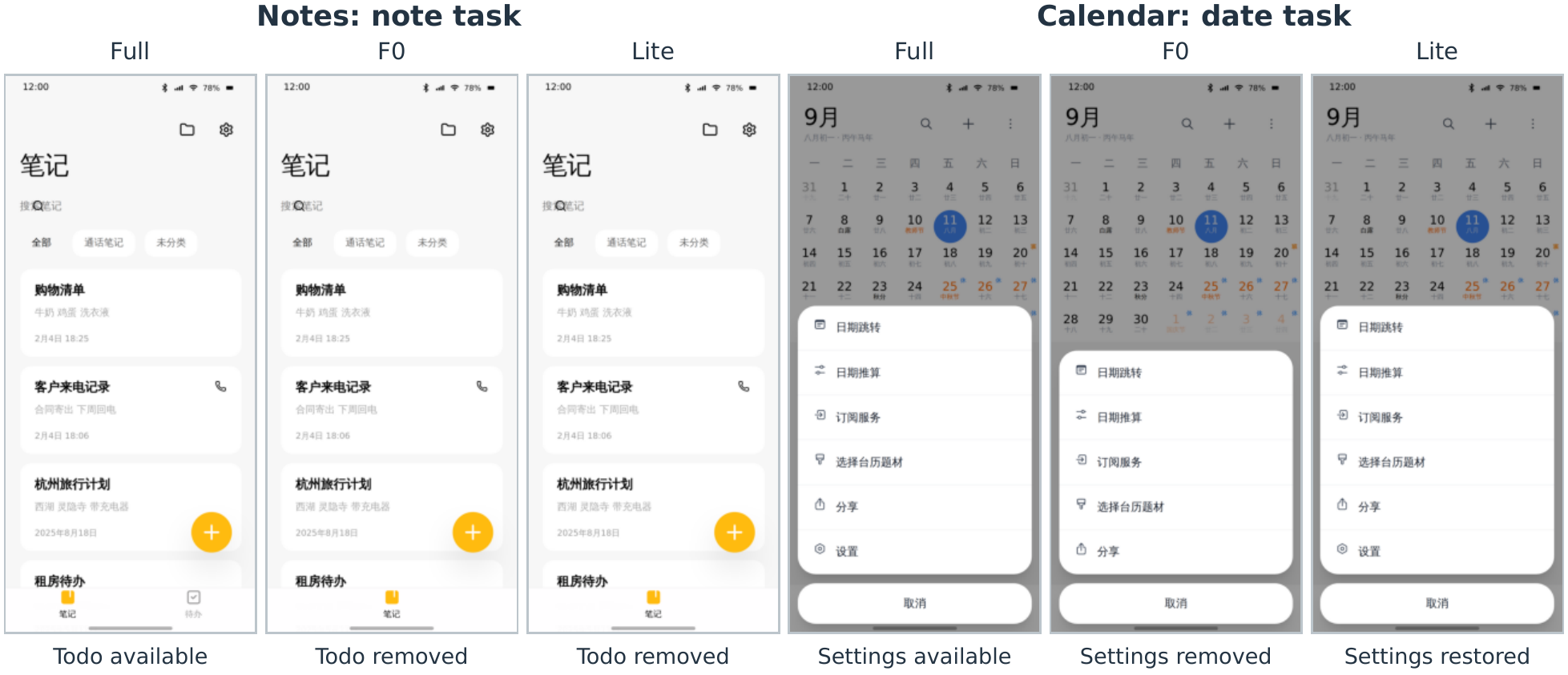}
 \caption{\textbf{Additional examples of functional pruning.}
 Notes loses its Todo entrance in both F0 and Lite for note tasks.
 In the Calendar date-tool profile, F0 removes Calendar settings and Lite
 restores them. Each group shows the same initialized state across modes.}
 \label{fig:extra-screens}
\end{figure*}

The nine-view panel spans the function profiles listed in \cref{tab:ui}.
All modes use the same task state, clock, and frontend.
We render a 360$\times$800 CSS-pixel viewport at device scale factor three,
waiting for fonts and two animation frames after opening each view.
\Cref{tab:ui} gives the counts, and \cref{fig:extra-screens} extends the main
paper's Clock and WeChat examples to Notes and Calendar.
\Cref{tab:complexity} summarizes functional coverage and the panel totals.
The main figure preserves the full viewport and circles the relevant controls.
For a removed Moments entry, a dashed circle marks the boundary between the
remaining rows. Measurements use the original screenshots without annotations.

\begin{table}[t]
 \centering
 \caption{\textbf{Functional complexity.} Disablements count template--module
 pairs. Visible control nodes are measured on nine fixed screens.}
 \label{tab:complexity}
 \small
 \input{tables/complexity}
\end{table}

\begin{table}[t]
 \centering
 \caption{\textbf{Visible control nodes on the fixed screen panel.}
 Switching function masks preserves application records in every measured view.}
 \label{tab:ui}
 \small
 \input{tables/ui_panel}
\end{table}

A counted control is a visible viewport-intersecting DOM node matching a
tagged action/entrance, button, button role, input, textarea, or select.
Multiple selector matches on one node count once, while a tagged wrapper and
an inner button count separately. Counts total 126 for Full, 113 for F0,
and 117 for Lite, a 7.14\% reduction from Full to Lite on this panel.
The measurement describes these selected views. Relative to the original
Full environment, Lite reduces the application source a pruned task needs
by 14.05\% of its lines on average over the 72 pruned templates, and it
keeps all 192 training templates.

To measure construction cost beyond source lines, we had the Codex coding
agent rebuild ten \gym{} applications, including WeChat and Bilibili, in
two styles. The complete style follows \gym{} and implements each
application as fully as possible. The pruned style follows Lite and
implements only the functions that the main task templates use, close to
F0. The pruned rebuild consumed about 33\% fewer tokens than the complete
one.

%% file: tables/complexity.tex
\begin{tabular}{@{}lrrr@{}}
\toprule
Quantity & Full & F0 & Lite \\
\midrule
Training templates & 192 & 192 & 192 \\
Modified templates & 0 & 78 & 72 \\
Disablements & 0 & 127 & 87 \\
Visible nodes (9 screens) & 126 & 113 & 117 \\
\bottomrule
\end{tabular}

%% file: tables/ui_panel.tex
\begin{tabular}{@{}lrrr@{}}
\toprule
Selected screen & Full & F0 & Lite \\
\midrule
Clock alarm & 23 & 20 & 21 \\
Clock world & 10 & 8 & 8 \\
Notes note & 13 & 12 & 12 \\
Notes todo & 16 & 15 & 15 \\
Calendar event menu & 17 & 16 & 16 \\
Calendar date menu & 17 & 16 & 17 \\
WeChat Me & 9 & 7 & 8 \\
WeChat settings & 12 & 11 & 12 \\
WeChat Discover & 9 & 8 & 8 \\
\bottomrule
\end{tabular}

%% file: sections/appendix/e_weighting.tex
\section{Episode Weights and Capability Retention}
\label{app:weighting}
\subsection{Balanced Episode Weights}
Let $T_i$ be the valid response count of episode $i$ and $\ell_{it}$ its
$t$-th response's token-averaged GSPO objective. Terminal advantage $A_i$
is broadcast to every response, and the default objective averages over all
valid responses in the batch.

Let $S=\sum_i T_i$ and $N$ be the number of episodes with valid responses.
With $C=S/N$ and $w_i=C/T_i$, the weighted objective is
\begin{align}
 \mathcal{L}_{\mathrm{episode}}
 &=\frac{1}{S}\sum_i\sum_{t=1}^{T_i}\frac{S}{NT_i}\ell_{it}\nonumber\\
 &=\frac{1}{N}\sum_i\frac{1}{T_i}\sum_{t=1}^{T_i}\ell_{it}.
 \label{eq:app-objective}
\end{align}
The mean weight across valid responses is
\begin{equation}
 \frac{1}{S}\sum_i T_iw_i
 =\frac{1}{S}\sum_i C=1.
 \label{eq:app-mean}
\end{equation}
Thus an episode's raw advantage coefficient changes from $T_iA_i$ to
$CA_i$. Episode length no longer multiplies this coefficient. Because the
weights are positive, scaling the advantage scales the response's clipped
surrogate by the same factor. Token gradients and minibatch updates still
depend on the responses that were generated.

The implementation counts nonempty response rows by trajectory and episode
identity, excluding padding. After terminal group normalization and
broadcasting, it multiplies each row's advantage and return by $C/T_i$, while response-token averaging and minibatching stay unchanged.
The recorded mean live weight differs from one by less than
$2\times10^{-6}$. Implementation checks cover single-turn episodes,
reordering, padding, and the gradients of the actual clipped loss.

\subsection{Full Results and Paired Retention}
Each arm begins from the same F0 step~12 model and optimizer state and trains
the original remaining 48 templates. The common checkpoint is freshly
evaluated alongside the four endpoints, yielding 224/465 overall and
81/153 retention-target successes. All retention transitions below use this
common evaluation. The complete 2325-episode evaluation has zero execution
or judging errors.

\begin{table}[t]
 \centering
 \caption{\textbf{Complete results of the environment--weighting experiment.}
 Denominators are 153 (Ret.), 183 (Aff.), 213 (Unmod.), 69 (OOD), and
 465 (All). Retention targets are a subset of unmodified ID. The environment
 and weighting columns describe the four training tails from the step-12 start.}
 \label{tab:allcohorts}
 \small
 \setlength{\tabcolsep}{2.5pt}
 \input{tables/all_cohorts}
\end{table}

Paired transitions distinguish retained capability from newly solved
instances (\cref{tab:retention}). Under response averaging, F0 retains
66 initial successes and gains 14, whereas Full retains 57 and gains five.
The largest difference in lost-success outcomes is looping: 21 for Full
and seven for F0. Balancing changes Full's retained count to 69 and F0's
to 61. These transitions explain why the final environment gap reverses
under a change to episode weighting.

\begin{table}[t]
 \centering
 \caption{\textbf{Transitions on the primary retention set.} The common
 start has 81 successes and 72 failures. Kept plus new gives the endpoint success count, and Loops counts initial successes ending in loops.}
 \label{tab:retention}
 \small
 \setlength{\tabcolsep}{3pt}
 \input{tables/retention}
\end{table}

On the retention set, the F0--Full gap is +11.76 points under response
averaging and $-6.54$ under balancing. Their interaction is +18.30 points.
Native-goal success gives an interaction of +18.95 points:
Full/F0 goal counts are 65/82 under response averaging
and 84/72 under balancing. Both measures exhibit the same directional change.

\subsection{Update Contributions}
\Cref{tab:train} summarizes the generated training episodes.
Response-averaged Full and F0 have nearly identical success counts,
142 and 141 of 384, but F0 produces more GUI responses.
On the 19 affected templates alone, Full/F0 have 57/60 successes and
1734/1915 responses. Thus the retention advantage accompanies changes in
the distribution of trajectories even when aggregate training success is
similar.

\begin{table}[t]
 \centering
 \caption{\textbf{Realized training over steps~13--16.}
 Success counts are out of 384 episodes, and responses exclude padding.
 Adam denotes added optimizer updates, and Format denotes output-format
 failures.}
 \label{tab:train}
 \small
 \setlength{\tabcolsep}{3pt}
 \input{tables/training}
\end{table}

We measure positive coefficient exposure as
$E^+=\sum_{i:A_i>0}T_iA_i$ and negative exposure magnitude as
$E^-=\sum_{i:A_i<0}T_i|A_i|$ on the affected tail templates.
Response averaging gives $E^+=386.2/320.3$ and $E^-=416.3/500.8$ for
Full/F0. Balancing replaces $T_i$ by $C$, and its positive and negative
magnitudes are 431.7 for Full and 464.8 for F0. The symmetry under balancing
follows from group-normalized advantages and equal episode coefficients.
These scalar measurements show which advantages are repeated more often, but their corresponding parameter updates also depend on token-level gradients.

%% file: tables/all_cohorts.tex
\begin{tabular}{@{}llrrrrr@{}}
\toprule
Tail env. & Weighting & Ret. & Aff. & Unmod. & OOD & All \\
\midrule
Step-12 start & -- & 81 & 83 & 105 & 36 & 224 \\
Full & Response & 62 & 71 & 89 & 31 & 191 \\
F0 & Response & 80 & 85 & 106 & 43 & 234 \\
Full & Episode & 80 & 79 & 105 & 37 & 221 \\
F0 & Episode & 70 & 82 & 97 & 40 & 219 \\
\bottomrule
\end{tabular}

%% file: tables/retention.tex
\begin{tabular}{@{}llrrrr@{}}
\toprule
Tail env. & Weighting & Kept /81 & New & Lost & Loops \\
\midrule
Full & Response & 57 & 5 & 24 & 21 \\
F0 & Response & 66 & 14 & 15 & 7 \\
Full & Episode & 69 & 11 & 12 & 4 \\
F0 & Episode & 61 & 9 & 20 & 7 \\
\bottomrule
\end{tabular}

%% file: tables/training.tex
\begin{tabular}{@{}llrrrr@{}}
\toprule
Tail env. & Weighting & Success & Responses & Adam & Format \\
\midrule
Full & Response & 142 & 3997 & 44 & 3 \\
F0 & Response & 141 & 4196 & 45 & 0 \\
Full & Episode & 149 & 4249 & 47 & 2 \\
F0 & Episode & 141 & 4274 & 46 & 1 \\
\bottomrule
\end{tabular}

%% file: sections/appendix/f_alternatives.tex
\section{Additional Mechanism Checks}
\label{app:alternatives}
\subsection{Pruned Evaluation Context}
We test whether F0 prefers its training context by crossing fixed Full/F0
policies with Full/F0 evaluation environments. The comparison covers 61 affected
templates and 12 unmodified controls at three seeds, for 876
episodes.

The affected-task results give a negative context-adaptation interaction.
The Full policy succeeds on 101/183 instances in Full and 104/183 in F0 context, and the F0 policy succeeds on 102/183 and 98/183, respectively.
The interaction is $-7/183=-3.83$ points.
Four outcome flips also occur on the unmodified controls,
where the context switch changes no functions. On the affected tasks,
switching to F0 context adds three successes for the Full policy and
removes four for the F0 policy.

\subsection{Source-Only Transfer}
A second check asks whether practicing pruned source tasks is sufficient
to improve other tasks before direct target practice. We train a Full/F0
pair on 72 source templates and then add common practice on 24 targets.
Evaluations along this sequence contain 1410 episodes in total.

Full has higher target success at both source-only checkpoints.
Before target practice, F0 succeeds on 73/213 primary instances and Full
on 81/213. After further source training, the corresponding counts are
89/213 and 95/213. The Full lead decreases from eight to six successes
as source training continues.

%% file: sections/appendix/g_cases.tex
\section{Paired Behavioral Examples}
\label{app:cases}
We illustrate both signs of the environment--weighting interaction.
Among the primary templates, we select the largest positive and largest
negative interactions, break ties lexically, and take the first task seed
with the corresponding sign. \Cref{tab:cases} summarizes the resulting
two seed~42 instances.

\begin{table}[t]
 \centering
 \caption{\textbf{Paired outcomes on two retention tasks.}
 Entries give outcome and action count. The two examples have opposite
 template-level interactions.}
 \label{tab:cases}
 \small
 \setlength{\tabcolsep}{3pt}
 \input{tables/cases}
\end{table}

\paragraph{Ticket-price query.}
The Railway task asks for a ticket price on a specified date, and the answer is 58.
Response-averaged F0 and balanced Full open Orders, inspect the user's
tickets, and answer in three actions. Response-averaged Full and balanced
F0 instead open the date selector and repeat ineffective swipes until the
nine-action loop stop. The step-12 starting policy also fails, repeatedly selecting
an order entrance from the wrong page. The successful endpoints discover
a usable information-retrieval route on a task absent from the training tail.

\paragraph{Profile visibility.}
The Reading task asks to make a profile visible to mutual followers.
Response-averaged F0 misses the settings
route and reaches the 30-action limit. The other three endpoints reach
Settings, Privacy, and Profile Visibility, change the requested setting,
and complete the task in 6--10 actions. This example shows that the
retention benefit varies across tasks even within the same environment--weighting comparison.

%% file: tables/cases.tex
\begin{tabular}{@{}llll@{}}
\toprule
Environment & Weighting & Ticket price & Profile visibility \\
\midrule
Step-12 start & -- & Loop, 9 & Success, 8 \\
Full & Response & Loop, 9 & Success, 6 \\
F0 & Response & Success, 3 & Limit, 30 \\
Full & Episode & Success, 3 & Success, 7 \\
F0 & Episode & Loop, 9 & Success, 10 \\
\bottomrule
\end{tabular}

%% file: references.bib
@inproceedings{wu2026mobilegym,
  title = {{MobileGym: A Verifiable and Highly Parallel Simulation Platform for Mobile GUI Agent Research}},
  author = {Wu, Dingbang and Hao, Rui and Wang, Haiyang and others},
  booktitle = {Proceedings of the 2026 Conference on Empirical Methods in Natural Language Processing},
  pages = {23223--23247},
  publisher = {Association for Computational Linguistics},
  year = {2026},
  doi = {10.18653/2026.emnlp-main.1009},
  url = {https://aclanthology.org/2026.emnlp-main.1009/}
}

@article{cao2026guigenesis,
  title = {{{GUI-GENESIS}: Automated Synthesis of Efficient Environments with Verifiable Rewards for GUI Agent Post-Training}},
  author = {Cao, Yuan and Ran, Dezhi and Wu, Mengzhou and others},
  journal = {arXiv preprint arXiv:2602.14093},
  year = {2026},
  url = {https://arxiv.org/abs/2602.14093}
}

@article{bai2025qwen3vl,
  title = {{{Qwen3-VL} Technical Report}},
  author = {Bai, Shuai and Cai, Yuxuan and Chen, Ruizhe and others},
  journal = {arXiv preprint arXiv:2511.21631},
  year = {2025},
  url = {https://arxiv.org/abs/2511.21631}
}

@inproceedings{zhou2024webarena,
  title = {{{WebArena}: A Realistic Web Environment for Building Autonomous Agents}},
  author = {Zhou, Shuyan and Xu, Frank F. and Zhu, Hao and others},
  booktitle = {International Conference on Learning Representations},
  year = {2024},
  url = {https://openreview.net/forum?id=oKn9c6ytLx}
}

@inproceedings{koh2024visualwebarena,
  title = {{VisualWebArena: Evaluating Multimodal Agents on Realistic Visually Grounded Web Tasks}},
  author = {Koh, Jing Yu and Lo, Robert and Jang, Lawrence and others},
  booktitle = {Proceedings of the 62nd Annual Meeting of the Association for Computational Linguistics (Volume 1: Long Papers)},
  pages = {881--905},
  publisher = {Association for Computational Linguistics},
  year = {2024},
  doi = {10.18653/v1/2024.acl-long.50},
  url = {https://aclanthology.org/2024.acl-long.50/}
}

@inproceedings{rawles2025androidworld,
  title = {{{AndroidWorld}: A Dynamic Benchmarking Environment for Autonomous Agents}},
  author = {Rawles, Christopher and Clinckemaillie, Sarah and Chang, Yifan and others},
  booktitle = {International Conference on Learning Representations},
  year = {2025},
  url = {https://openreview.net/forum?id=il5yUQsrjC}
}

@inproceedings{xie2024osworld,
  title = {{{OSWorld}: Benchmarking Multimodal Agents for Open-Ended Tasks in Real Computer Environments}},
  author = {Xie, Tianbao and Zhang, Danyang and Chen, Jixuan and others},
  booktitle = {Advances in Neural Information Processing Systems},
  volume = {37},
  pages = {52040--52094},
  publisher = {Curran Associates, Inc.},
  year = {2024},
  doi = {10.52202/079017-1650},
  url = {https://proceedings.neurips.cc/paper_files/paper/2024/file/5d413e48f84dc61244b6be550f1cd8f5-Paper-Datasets_and_Benchmarks_Track.pdf}
}

@inproceedings{sun2025osgenesis,
  title = {{{OS-Genesis}: Automating GUI Agent Trajectory Construction via Reverse Task Synthesis}},
  author = {Sun, Qiushi and Cheng, Kanzhi and Ding, Zichen and others},
  booktitle = {Proceedings of the 63rd Annual Meeting of the {A}ssociation for {C}omputational {L}inguistics (Volume 1: Long Papers)},
  pages = {5555--5579},
  publisher = {Association for Computational Linguistics},
  year = {2025},
  doi = {10.18653/v1/2025.acl-long.277},
  url = {https://aclanthology.org/2025.acl-long.277/}
}

@article{shao2024deepseekmath,
  title = {{{DeepSeekMath}: Pushing the Limits of Mathematical Reasoning in Open Language Models}},
  author = {Shao, Zhihong and Wang, Peiyi and Zhu, Qihao and others},
  journal = {arXiv preprint arXiv:2402.03300},
  year = {2024},
  url = {https://arxiv.org/abs/2402.03300}
}

@article{zheng2025gspo,
  title = {{Group Sequence Policy Optimization}},
  author = {Zheng, Chujie and Liu, Shixuan and Li, Mingze and others},
  journal = {arXiv preprint arXiv:2507.18071},
  year = {2025},
  url = {https://arxiv.org/abs/2507.18071}
}

@inproceedings{liu2025r1zero,
  title = {{Understanding {R1-Zero-Like} Training: A Critical Perspective}},
  author = {Liu, Zichen and Chen, Changyu and Li, Wenjun and others},
  booktitle = {Conference on Language Modeling},
  year = {2025},
  url = {https://openreview.net/forum?id=5PAF7PAY2Y}
}

@inproceedings{bai2024digirl,
  title = {{{DigiRL}: Training In-The-Wild Device-Control Agents with Autonomous Reinforcement Learning}},
  author = {Bai, Hao and Zhou, Yifei and Cemri, Mert and others},
  booktitle = {Advances in Neural Information Processing Systems},
  volume = {37},
  pages = {12461--12495},
  publisher = {Curran Associates, Inc.},
  year = {2024},
  doi = {10.52202/079017-0397},
  url = {https://proceedings.neurips.cc/paper_files/paper/2024/file/1704ddd0bb89f159dfe609b32c889995-Paper-Conference.pdf}
}

@inproceedings{qi2025webrl,
  title = {{{WebRL}: Training {LLM} Web Agents via Self-Evolving Online Curriculum Reinforcement Learning}},
  author = {Qi, Zehan and Liu, Xiao and Iong, Iat Long and others},
  booktitle = {International Conference on Learning Representations},
  year = {2025},
  url = {https://openreview.net/forum?id=oVKEAFjEqv}
}

@inproceedings{zhang2024llamatouch,
  title = {{{LlamaTouch}: A Faithful and Scalable Testbed for Mobile {UI} Task Automation}},
  author = {Zhang, Li and Wang, Shihe and Jia, Xianqing and others},
  booktitle = {Proceedings of the 37th Annual ACM Symposium on User Interface Software and Technology},
  pages = {1--13},
  publisher = {Association for Computing Machinery},
  year = {2024},
  doi = {10.1145/3654777.3676382}
}

@article{im2025mobibench,
  title = {{{MobiBench}: Multi-Branch, Modular Benchmark for Mobile {GUI} Agents}},
  author = {Im, Youngmin and Jo, Byeongung and Wi, Jaeyoung and others},
  journal = {arXiv preprint arXiv:2512.12634},
  year = {2025},
  url = {https://arxiv.org/abs/2512.12634}
}

@article{lu2025uir1,
  title = {{{UI-R1}: Enhancing Efficient Action Prediction of {GUI} Agents by Reinforcement Learning}},
  author = {Lu, Zhengxi and Chai, Yuxiang and Guo, Yaxuan and others},
  journal = {arXiv preprint arXiv:2503.21620},
  year = {2025},
  url = {https://arxiv.org/abs/2503.21620}
}

@article{liu2026scalewob,
  title = {{{ScaleWoB}: Guiding {GUI} Agents with Coding Agents via Large-Scale Environmental Synthesis}},
  author = {Liu, Guohong and Ye, Jialei and Gao, Pengzhi and others},
  journal = {arXiv preprint arXiv:2605.25160},
  year = {2026},
  url = {https://arxiv.org/abs/2605.25160}
}

@article{toyama2021androidenv,
  title = {{{AndroidEnv}: A Reinforcement Learning Platform for {Android}}},
  author = {Toyama, Daniel and Hamel, Philippe and Gergely, Anita and others},
  journal = {arXiv preprint arXiv:2105.13231},
  year = {2021},
  url = {https://arxiv.org/abs/2105.13231}
}

@inproceedings{xu2025androidlab,
  title = {{{AndroidLab}: Training and Systematic Benchmarking of {Android} Autonomous Agents}},
  author = {Xu, Yifan and Liu, Xiao and Sun, Xueqiao and others},
  booktitle = {Proceedings of the 63rd Annual Meeting of the {A}ssociation for {C}omputational {L}inguistics (Volume 1: Long Papers)},
  pages = {2144--2166},
  publisher = {Association for Computational Linguistics},
  year = {2025},
  doi = {10.18653/v1/2025.acl-long.107},
  url = {https://aclanthology.org/2025.acl-long.107/}
}

@article{tong2026maigym,
  title = {{{MAI-Gym}: An Open All-in-One Platform for {GUI} Agent Evaluation and {RL} Training across Mobile and Computer Environments}},
  author = {Tong, Panrong and Kong, Quyu and Li, Pengxiang and others},
  journal = {arXiv preprint arXiv:2610.20939},
  year = {2026},
  url = {https://arxiv.org/abs/2610.20939}
}

@article{wang2025uisimulator,
  title = {{{LLMs} as Scalable, General-Purpose Simulators For Evolving Digital Agent Training}},
  author = {Wang, Yiming and Yin, Da and Cui, Yuedong and others},
  journal = {arXiv preprint arXiv:2510.14969},
  year = {2025},
  url = {https://arxiv.org/abs/2510.14969}
}

@article{tang2026phonebuddy,
  title = {{{PhoneBuddy}: Training Open Models for Agentic Phone Use}},
  author = {Tang, Zhengyang and Lai, Xin and Lyu, Pengyuan and others},
  journal = {arXiv preprint arXiv:2606.23049},
  year = {2026},
  url = {https://arxiv.org/abs/2606.23049}
}

@article{wang2026stamp,
  title = {{{STAMP}: Training Explicit Memory for Mobile {GUI} Agents in Controllable and Scalable Virtual Environments}},
  author = {Wang, Junyang and Xu, Haiyang and Zhang, Xi and others},
  journal = {arXiv preprint arXiv:2605.29324},
  year = {2026},
  url = {https://arxiv.org/abs/2605.29324}
}

@article{xu2026mobileagentv35,
  title = {{{Mobile-Agent-v3.5}: Multi-platform Fundamental {GUI} Agents}},
  author = {Xu, Haiyang and Zhang, Xi and Liu, Haowei and others},
  journal = {arXiv preprint arXiv:2602.16855},
  year = {2026},
  url = {https://arxiv.org/abs/2602.16855}
}

@inproceedings{wang2025distrl,
  title = {{{DistRL}: An Asynchronous Distributed Reinforcement Learning Framework for On-Device Control Agent}},
  author = {Wang, Taiyi and Wu, Zhihao and Liu, Jianheng and others},
  booktitle = {International Conference on Learning Representations},
  year = {2025},
  url = {https://openreview.net/forum?id=LPG8pPSfQD}
}

@inproceedings{xu2026mobilerl,
  title = {{{MobileRL}: Online Agentic Reinforcement Learning for Mobile {GUI} Agents}},
  author = {Xu, Yifan and Liu, Xiao and Liu, Xinghan and others},
  booktitle = {International Conference on Learning Representations},
  year = {2026},
  url = {https://openreview.net/forum?id=C3F0G9nXhl}
}

@article{yang2025zerogui,
  title = {{{ZeroGUI}: Automating Online {GUI} Learning at Zero Human Cost}},
  author = {Yang, Chenyu and Su, Shiqian and Liu, Shi and others},
  journal = {arXiv preprint arXiv:2505.23762},
  year = {2025},
  url = {https://arxiv.org/abs/2505.23762}
}

@article{wang2025uitars2,
  title = {{{UI-TARS-2} Technical Report: Advancing {GUI} Agent with Multi-Turn Reinforcement Learning}},
  author = {Wang, Haoming and Zou, Haoyang and Song, Huatong and others},
  journal = {arXiv preprint arXiv:2509.02544},
  year = {2025},
  url = {https://arxiv.org/abs/2509.02544}
}

@inproceedings{zhang2026gsar,
  title = {{{GSAR}: Goal-State-Anchor Rewards for Mobile {GUI} Agents with Self-Evolving Data Synthesis}},
  author = {Zhang, Long and Chen, Yuhan and Zhang, Chaoran and others},
  booktitle = {Proceedings of the 2026 Conference on Empirical Methods in Natural Language Processing},
  pages = {47714--47730},
  publisher = {Association for Computational Linguistics},
  year = {2026},
  doi = {10.18653/2026.emnlp-main.2095},
  url = {https://aclanthology.org/2026.emnlp-main.2095/}
}

@article{wang2026vericua,
  title = {{{VeriCUA}: Learning Proactive Verification for Reliable {GUI} Agent Rewards}},
  author = {Wang, Zhizhou and Lu, Yaxi and Zhang, Zhong and others},
  journal = {arXiv preprint arXiv:2610.26469},
  year = {2026},
  url = {https://arxiv.org/abs/2610.26469}
}

@inproceedings{lee2025bmoca,
  title = {{Benchmarking Mobile Device Control Agents across Diverse Configurations}},
  author = {Lee, Juyong and Min, Taywon and An, Minyong and others},
  booktitle = {Proceedings of the 4th Conference on Lifelong Learning Agents},
  series = {Proceedings of Machine Learning Research},
  volume = {330},
  pages = {261--306},
  publisher = {PMLR},
  year = {2025},
  url = {https://proceedings.mlr.press/v330/lee26a.html}
}

@article{cui2026agenticrm,
  title = {{Agentic Reward Modeling: Verifying {GUI} Agent via Progressive Trajectory-Grounded Interaction}},
  author = {Cui, Chaoqun and Huang, Jing and Wang, Shijing and others},
  journal = {arXiv preprint arXiv:2602.00575},
  year = {2026},
  url = {https://arxiv.org/abs/2602.00575}
}

@inproceedings{li2024androidcontrol,
  title = {{On the Effects of Data Scale on {UI} Control Agents}},
  author = {Li, Wei and Bishop, William and Li, Alice and others},
  booktitle = {Advances in Neural Information Processing Systems},
  volume = {37},
  pages = {92130--92154},
  publisher = {Curran Associates, Inc.},
  year = {2024},
  doi = {10.52202/079017-2925}
}
